\documentclass[runningheads]{llncs}
\usepackage[T1]{fontenc}
\usepackage{graphicx} % for \resizebox
\usepackage{booktabs} % for \toprule, \midrule, and \bottomrule
\usepackage{subcaption}

\usepackage{multirow}

\usepackage{graphicx}
\usepackage{float}
\usepackage{amsmath}
\begin{document}
\title{Aspect-Based Few-Shot Learning}
\author{Tim van Engeland\inst{1}, Lu Yin\inst{1,2} \and
Vlado Menkovski\inst{1}}

%\and
%Third Author\inst{3}\orcidID{2222--3333-4444-5555}}
%
\authorrunning{T. van Engeland et al.}
\institute{Eindhoven University of Technology, Eindhoven, The Netherlands \and
University of Surrey, Guildford, United Kingdom
}
\maketitle              % typeset the header of the contribution
\begin{abstract}
We generalize the formulation of few-shot learning by introducing the concept of an aspect. In the traditional formulation of few-shot learning, there is an underlying assumption that a single "true" label defines the content of each data point. This label serves as a basis for the comparison between the query object and the objects in the support set. However, when a human expert is asked to execute the same task without a predefined set of labels, they typically consider the rest of the data points in the support set as context. This context specifies the level of abstraction and the aspect from which the comparison can be made. In this work, we introduce a novel architecture and training procedure that develops a context given the query and support set and implements aspect-based few-shot learning that is not limited to a predetermined set of classes. We demonstrate that our method is capable of forming and using an aspect for few-shot learning on the Geometric Shapes and Sprites dataset. The results validate the feasibility of our approach compared to traditional few-shot learning.

\end{abstract}
\section{Introduction}\label{chapter:introduction} 
Machine learning (ML) methods have been highly successful in image analysis tasks such as image classification \cite{he2016deep}. Nevertheless, this formulation poses a significant limitation for the more general goal of recognizing the content of an image or other high-dimensional data. Specifically, to precisely recognize the content of such data, many more labels are typically required. Such labels may not be available during training, may form various hierarchical relationships, the language they use may not be exact, and depending on the context these labels may vary for a given data point. In contrast to this, humans excel at such recognition tasks. We can understand the content of the image data and adjust this understanding to different aspects.  

The few-shot learning (FSL) approaches generalize over the classification formulation, as the model is no longer limited to a predefined set of classes, but can generalize from a few training examples \cite{parnami2022learning} to recognize image content beyond what was available during training. The objective of the FSL model is to learn the transferable knowledge from the "base" data (training data) with a small amount of labeled data and to be able to apply it to a "novel" dataset. 
%Furthermore, FSL is particularly suitable for settings where the ratio of available labels per class is low. 

\begin{figure}[H]
    \begin{subfigure}{\textwidth}
        \centering
        \includegraphics[width=0.8\linewidth]{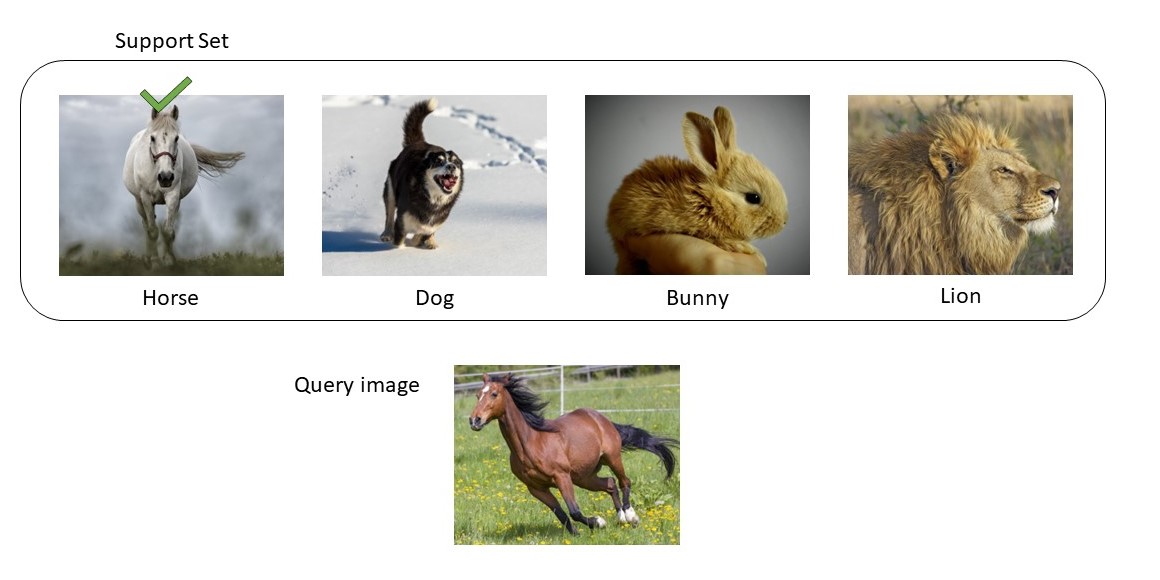}
        \caption{FSL classification based on the predefined label will assign the correct label “horse” to the query image.}
        \label{fig:classic}
    \end{subfigure}
    \begin{subfigure}{\textwidth}
        \centering
        \includegraphics[width=0.8\linewidth]{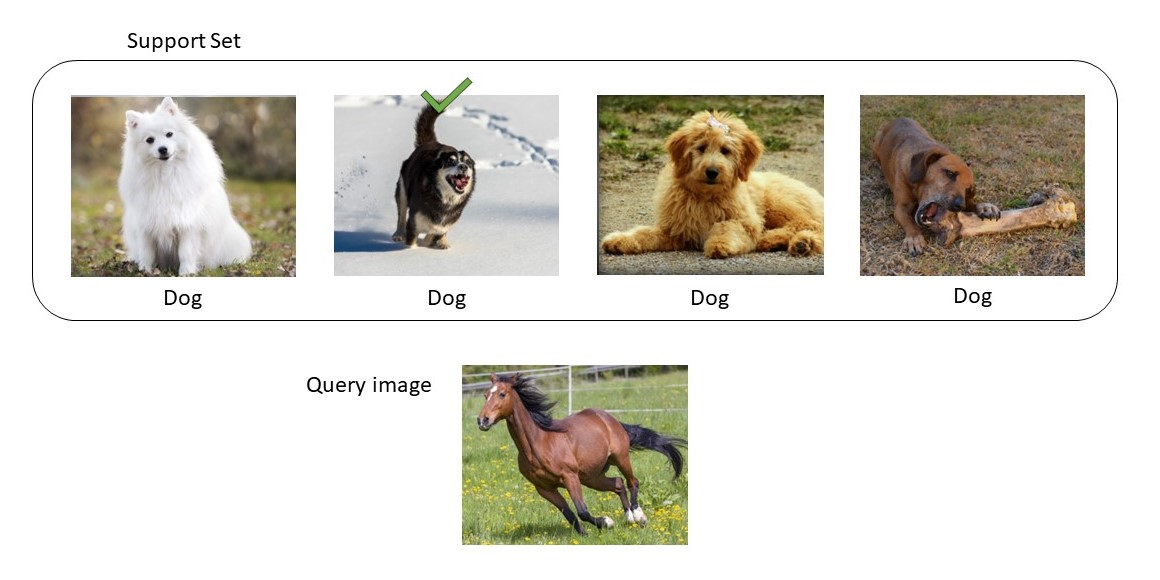}
        \caption{Each image in the support set has the label "dog", while the query image is labeled as "horse". Therefore, there is no match for the query image. However, given the support set if we change the aspect from species to physical activity, the query image can be matched based on the label "running".}
        \label{fig:running}
    \end{subfigure}
    \begin{subfigure}{\textwidth}
        \centering
        \includegraphics[width=0.8\linewidth]{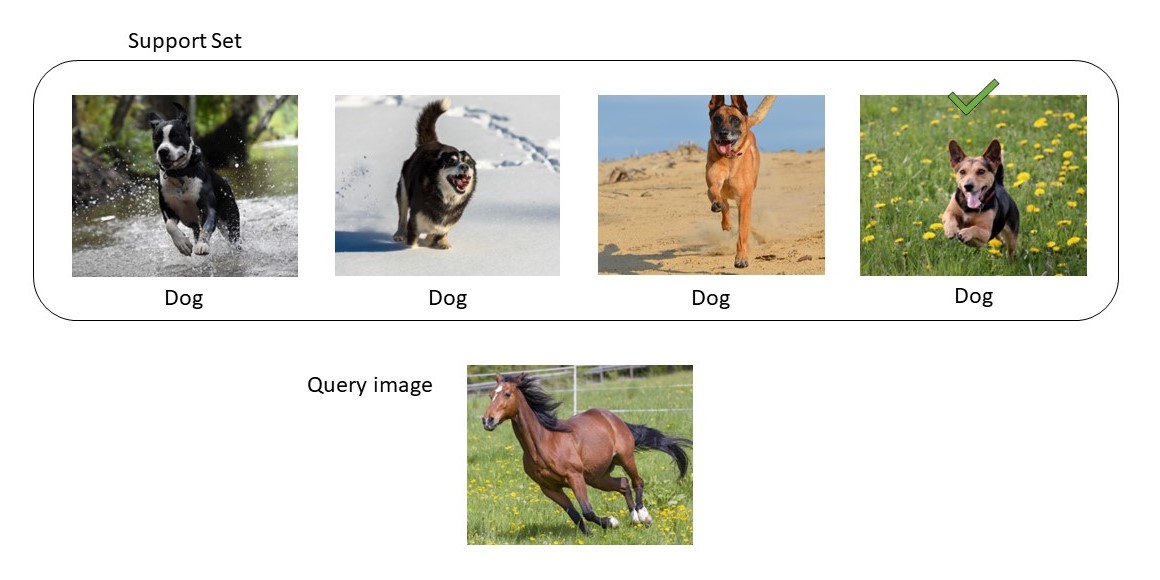}
        \caption{All the images in the support set are running dogs, but the background of the images is different. If this aspect is considered, a match can be made as only the fourth image shares a common background of a grass field with flowers with the query image.}
        \label{fig:field}
    \end{subfigure}    
    \caption{Examples of different FSL cases that illustrate how taking different aspects results with different matching between the query and support set data points}
    \label{fig:fsl-contexts}
\end{figure}

%The model is presented with two parts to gain the knowledge: a query instance and a support set. The model has to classify the query instance based on the examples of the support set. By repeating the task on the 'base' data, it will eventually learn how to apply the same principles on the same task but with a different dataset.

There have been multiple developments of the base formulation of FSL\cite{parnami2022learning}, however, these formulations are still limited to matching a datapoint to a unique label.  We demonstrate the shortcomings of this formulation with the following three examples of four-way-one-shot learning given in Figure~\ref{fig:fsl-contexts}. Figure \ref{fig:classic} shows a standard example FSL, there is one image of four distinct classes: "horse", "dog", "bunny", and "lion". In this example, we recognize that the query image is in line with the content of the first image and should therefore be labeled as a "horse". However, the second set of images in Figure \ref{fig:running}, show images of only dogs. Despite this, we can still say that the second image is the most similar to the query image. The label "dog" does not fully describe the content of the image. The first image contains a sitting dog, the second shows a running dog, in the third one the dog lays down, and in the last image, the dog is chewing on a bone. Given this context, we can discern that the physical activity of the dogs is different for each image. So, rather than the presence of the dog being the differentiating factor, we can deduce that the activity of running is a property that both the query and (only) the second image share. The third example in Figure \ref{fig:field} further demonstrates the point. Each image in the support set displays a running dog. However, each of them runs on a different surface. From left to right, the dogs run in the water, snow, sand and in a field. Therefore, the query image can be matched to (only) the fourth dog as they are both running on grass.  In every example, the second image in the support set is the same, as well as the query image, but the matching differs given the rest of the images in the support set. The first support set has images with varying animals. Therefore, the matching was performed on the basis of the species. In the second case, the species was not a factor of variation, but rather the physical activity. Lastly, in the third case, only the background was a factor that allowed us to match a single support set image with the query image. The examples show that the whole support set plays a defining role in how we compare each instance of the support set with the query image. 

The current state-of-the-art FSL methods cannot account for those relationships between the instances of the support set because it assumes a single label assignment to each datapoint on which the matching is implemented. As a consequence, only the first case given in Figure \ref{fig:running} would work with the exact match. For the second case we would need an additional set of labels based on the physical activity. However, adding multiple labels is not efficient and does not address this problem in general. 

In this work we propose an extension of the FSL formulation that incorporates an aspect that is formed by all the data instances in the support set. Rather than learning to match using a unique label, the model learns the differentiating feature of the support set. Hereby, we will define the \textit{aspect} as the varying feature(s) in the support set and implement the matching based on the values of that feature (which we refer to as properties of the data points). With these definitions in place, we present a framework that learns the aspect given by the support set and, as such, identifies the correct assignment between the query and the instances of the support set. 

\section{Related work}

Few-shot learning (FSL) is a broad research field that can be categorized into two main approaches: non-meta-learning and meta-learning~\cite{parnami2022learning}. Non-meta-learning focuses primarily on transfer learning~\cite{zhuang2020comprehensive}, where knowledge from a pre-trained model is transferred to a new task to alleviate the need to train from scratch when data is scarce. On the other hand, meta-learning, often referred to as ``learning to learn,'' has gained significant attention due to its potential to address few-shot learning problems. Meta-learning approaches can be divided into three categories. \textit{Metric-based methods} aim to learn a shared metric space where few-shot predictions can be effectively made by measuring similarities between data points. Notable examples include Prototypical Networks~\cite{snell2017prototypical}, which compute class prototypes in the embedding space; Relation Networks~\cite{sung2018learning}, which learn a deep distance metric for comparing query and support examples; and Matching Networks~\cite{vinyals2016matching}, which leverage attention mechanisms over embeddings of labeled examples. Furthermore, psychometric-based metric learning methods take advantage of human perception~\cite{yin2021hierarchical,yin2021knowledge,yin2021semantic} to learn the similarity based on psychometric testing. \textit{Optimization-based methods} focus on adapting gradient-based optimization algorithms to enable rapid learning on new tasks with limited data. Representative works include Model-Agnostic Meta-Learning (MAML)~\cite{finn2017model}, which learns model parameters that are sensitive to rapid adaptation; and Reptile~\cite{nichol2018reptile}, an algorithm that approximates MAML with first-order gradients. In this work, we focus on metric-based approaches for few-shot learning. A closely related study to our research is the Categorical Traversal Module (CTM)~\cite{li2019finding}. The CTM traverses the entire support set to identify task-relevant features based on both intra-class commonality and inter-class uniqueness. Consider a 2-way 4-shot task with the classes \textit{tiger} and \textit{cheetah}. Each tiger has its recognizable coat of reddish-orange with dark stripes (intra-class commonality) and similarly for cheetahs with their distinct coats. The coats are unique to their species (interclass uniqueness). Both animals belong to the Felidae family (cats) and share characteristics. Therefore, identifying task-relevant features is crucial for correctly distinguishing between a tiger and a cheetah. Nevertheless, as discussed in the introduction, these methods assume a single-class assignment and are unable to deal with the more general setting where class assignment is conditioned on the context given by the support set.

\section{Method}
\subsection{Aspect definition}
% 
% 1. Shortly re-iterate the limitations of few-shot learning (a mapping to a single label). We generalize it to mapping to another datapoint given a context. 
% 2. The mapping given a context is done on a specific aspect. The aspect is defined by the set of properties that allow for a unique mapping between the query image and only one of the support images.  
% 3. 
%
In a traditional few-shot classification approach, we are given a query set $\mathrm{Q}$ and a support set $\mathrm{S}$ of $\mathrm{n}$ distinct, previously unseen classes, with $\mathrm{k}$ examples each. 

The goal of the model is to learn to classify the query instance $x_q \in Q$, into one of the $\mathrm{n}$ classes defined by the elements of the support set $S$. In other words, there exists a data point $x_i \in \mathrm{S}$ that matches the query image $x_q$. 

We aim to generalize this class-based assignment by introducing the notion of an aspect. Specifically, rather than assuming that a single class describes the content of a datapoint, we assume that each datapoint is described by a set of properties $\mathrm{P}_i$. In the traditional setting, we can say that a class assignment also has a set of properties $\mathrm{P}_c$, such that if a data point is labeled with the particular class value, then the set of properties $\mathrm{P}_i$ is equal to the set of properties of the class $\mathrm{P}_c$. However, in the aspect-based few-shot learning formulation, there is no such assumption, and therefore the properties of the query $\mathrm{P}_q$ and the properties of the support set data points $\mathrm{P}_i$ may only have a partial overlap. 

In aspect-based few-shot learning, we formulate the matching task based on an aspect rather than a class. In contrast to the class, we define the aspect as a set of properties that are shared between the query and only one element of the support set. 

For a given element $x_m$ part of the support set $\mathrm{S}$ ($x_m \in \mathrm{S}$). The query $x_q$ matches $x_m$ if there exists an aspect $\mathrm{A}$ such that $\mathrm{A} \subset x_m$ and $\mathrm{A} \subset x_q$ and $\mathrm{A} \not \subset x_i$ for all $i$ where $i \not = m$.

Therefore, the aspect is not predetermined as the class, but rather becomes specified at query time as a combination of the properties of the query and support set elements.

\subsection{Deep Set Traversal Module}
\begin{figure}[!b]
    \centering
    \includegraphics[width=0.8\textwidth]{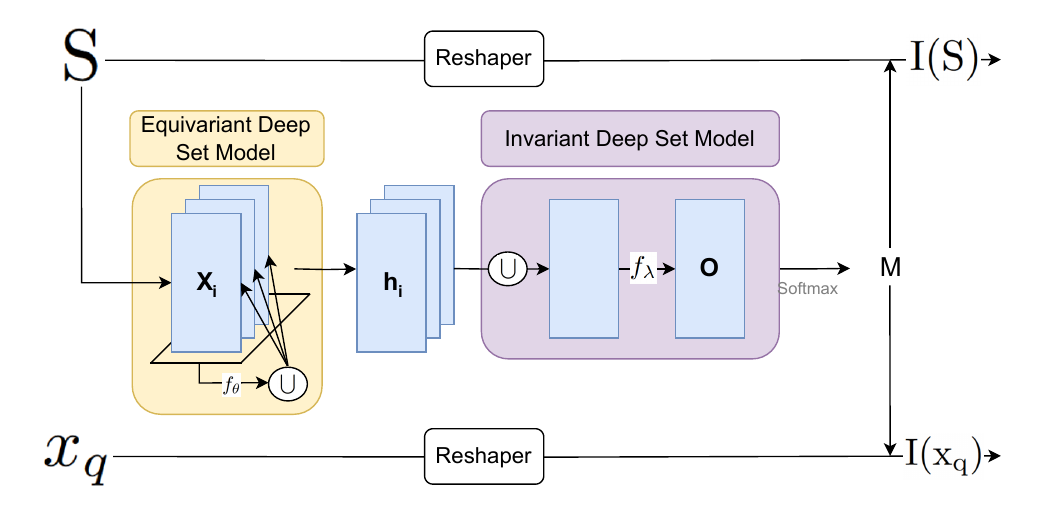}
    \caption{Detailed depiction of the DSTM components}
    \label{fig:DSTM}
\end{figure}

The CTM method discussed in the background section aligns with our goal of extracting the apsect from the support set, however, the model does not offer permutation invariance. As the order of the support set does not matter when comparing the support set to the query, the aspects need to be permutationally invariant. To address this, we present a modified architecture named Deep-Set Traversal Module (DSTM).
%Our proposed solution takes inspiration from CTM \cite{li2019finding}. However, the CTM is not permutationally invariant to the order of the data points in the support set, and as such it does not meet one of our requirements. Specifically in CTM, the data is concatenated along the channel axis before the projector and fed through the CNN module. Since a "general purpose" convolutional layer does not provide invariance over the channels, the order of examples in the support set affects the final embedding. To address this we present a modified architecture named Deep Set Traversal Module (DSTM) that is permutationally invariant to the order of the data in the support set.
\subsubsection{Permutation-equivariant Deep Set Model} In Figure \ref{fig:DSTM}, we replace the concentrator with a permutation-equivariant deep set model \cite{zaheer2017deep}. The purpose of the model is to enrich the embedding of each element with the information of the other members of the support set:

\begin{equation}\label{eq:equiDeepDSTM}
    h_i = f_{\theta}(x_i, \mathrm{P}_{\mathcal{N}(i)}) 
\end{equation}
\begin{equation}\label{eq:comNeighbour}
    \mathrm{P}_{\mathcal{N}(i)} = \bigcup_{j\in \mathcal{N}(i)}f_{\phi}(x_j)
\end{equation}
$\mathcal{N}(i)$ is the neighborhood of $x_i$, $f_\phi$ is the neighbor image embedding model.

This model extracts the aspect of each support set image by combining the image representation with the representation of the properties of the neighboring images. The model as specified in Equation \ref{eq:comNeighbour} extracts the shared properties in the neighbourhood $\mathrm{P}_{\mathcal{N}(i)}$ by individually processing the representation through embedding function $f_{\phi}$ and applying permutation-invariant operation $\bigcup$. The model as specified in equation \ref{eq:equiDeepDSTM} processes the set of properties $\mathrm{P}_{\mathcal{N}(i)}$ together with $x_i$ through the embedding function $f_{\theta}$ to determine the unique properties $\mathrm{h}_{i}$. This process occurs for each image $x_i$ in the support set $\mathrm{S}$.

%it processes the information of each embedding in $f(\mathrm{S}; \theta)$ individually through embedding function $g(;\mu)$. Afterward, we apply the permutation-invariant operation P over the neighboring embeddings. We denote the collection of permutation-invariant representations as $g_i(\mathrm{S})$. $g_i(\mathrm{S})$ is concatenated to the original embedding $f$($\mathrm{S}$; $\theta$) and put through a final embedding function $h(; \lambda)$, which leads to the final representation $h_i(\mathrm{S})$. Each embedding function can be a simple convolutional layer with ReLU activation or a ResNet block. Consider the output dimensions of $f(\mathrm{S};\theta)$ to be (N, $m_1, d_1, d_1$), where $m_1, d_1$ are the number of channels and width:

% \begin{equation*}
% \begin{split}
%     &f(\mathrm{S};\theta): \text{(N, $m_1, d_1, d_1$)}\xrightarrow[\text{}]{\text{CNN}} \text{(N, $m_2, d_1, d_1$)}\\
%     &g(\mathrm{S};\mu): \text{(N, $m_2, d_1, d_1$)} \xrightarrow[\text{neighbours}]{\text{P over}} \text{(N, $m_2, d_1, d_1$)}\\
%     &g_i(\mathrm{S}): \text{(N, $m_2, d_1, d_1$)}\xrightarrow[\text{concatenation}]{\text{with $f(\mathrm{S};\theta)$}} f(\mathrm{S};\theta),g_i(\mathrm{S}): \text{(N, $m_1+ m_2, d_1, d_1$)}  \xrightarrow[\text{}]{\text{CNN}} h_i(\mathrm{S}): \text{(N, $m_3, d_2, d_2$)}
% \end{split}
% \end{equation*}
\subsubsection{Permutation-invariant Deep Set Model} The second component is the permutation-invariant deep-set model \cite{zaheer2017deep}. In Equation \ref{eq:inDeepDSTM} we apply an additional operation $\bigcup$ on aspects $h_i$ to destroy the information regarding the order of the set. The permutation-invariant representation of the set is processed by another embedding function $f_{\lambda}$ to have a final output embedding $O$ representing the complete
set. As the last step, we apply a Softmax over the channel dimension of the output $O$ to create the mask $M$.

\begin{equation}\label{eq:inDeepDSTM}
    O = f_{\lambda}(\bigcup h_i)
\end{equation}

%The permutation-invariant representation of the set is processed by another neural network $k$(; $\eta$) to have a final output embedding representing the complete set. As the last step, we apply a Softmax over the channel dimension of the output of CNN as with the CTM:

%\begin{equation*}
%    \begin{split}
%        &h_i(\mathrm{S}): \text{(N, $m_3, d_2, d_2$)} \xrightarrow[\text{embeddings}]{\text{P over}}   P\left[ h_i(\mathrm{S})\right]: \text{(1, $m_3, d_2, d_2$)} \xrightarrow[\text{}]{\text{CNN}} O: \text{(1, $m_4, d_3, d_3$)}\\
%        &\xrightarrow[\text{}]{\text{Softax}} M: \text{(1, $m_4, d_3, d_3$)}
%    \end{split}
%\end{equation*}

\subsubsection{Reshaper} The purpose of the reshaper is to align the dimension of the feature map of the support set and the query with the dimensions of $M$

\subsubsection{Design choice in DSTM} The last step of the DSTM is to apply $M$ to the support set and query to be able to semantically match. This is done by an element-wise multiplication between the output permutation-variant deep set model $M$ and the reshaper output:
\begin{equation*}
    \begin{split}
        &\mathrm{I}(\mathrm{S}) =  M \odot r(\mathrm{S}) \\
        &\mathrm{I}(\mathrm{x_q}) =  M \odot r(\mathrm{x_q})
    \end{split}
\end{equation*}

\section{Evaluation}
%The experiments are designed to answer the following questions: (1) How to build a permutation invariant machine learning model for extracting aspects within a support set? (2) How to separate the training data from the test data for the task of aspect-based Few-Shot learning?

To evaluate this approach, our aim is to show that DSTM can deduce the aspect from the support set and use it to match the correct image. Based on our definition of an aspect, the matching is done by the unique discriminating property in the query and the support set. The existing FSL datasets and evaluation do not offer varying aspects as the matching is not done based not on different properties, but on the single class assignment assumption. We therefore set out to develop a novel evaluation specific to the aspect-based FSL method. 

\subsection{Data and setup}

\subsubsection{Data} 
To objectively assess the performance of our approach, we need fine control over the properties of each data point in the dataset such that we can have an unambiguous measure of accuracy. We achieve this by developing a fully synthetic dataset named \textit{geometric shapes}, where 4 factors are being varied to form a set of properties for each data point. We also repeat this evaluation approach on the \textit{sprite} public dataset. 

The \textit{geometric shapes} dataset includes two-dimensional geometric shapes with an image size of 112 by 112 pixels. Each image consists of a polygon with a hole in the middle. As such, a polygon can have a number of differentiating properties, namely: color, thickness, and pattern. The \textit{sprites} dataset is derived from the data generated to disentangle sequential data \cite{li2018disentangle}\cite{pixel2024art}. The available sprite sheets \cite{sprite2024} offer the possibility of creating custom characters. In this work, we are not interested in the complete sequence, but in individual frames. The selected frames represent different features: physical stance, body type, shirt, pants, and hair color.

\subsubsection{Support Sets}
A support set has several requirements to consider. First, the query image and the images of the support set do not share the same object type. That is, shapes in the \textit{geometric shapes} dataset and bodies in the \textit{sprites} dataset. We do not recognize objects as an aspect. Object recognition requires the simultaneous recognition of multiple properties at once. For example, a rhombus has several properties that define it as a rhombus. All sides of a rhombus are equal, diagonals bisect each other at 90$^{\circ}$, opposite sides are parallel in a rhombus, opposite angles are equal in a rhombus, and adjacent angles add up to 180$^{\circ}$. Additionally, we enforce the idea of no exact match between support and query by excluding objects as an aspect. 

Second, each support set must represent only a single discriminating property, and every other feature is a shared property. Altering multiple properties in a support set leads to ambiguity, while this evaluation requires a controlled environment without ambiguity. 

Third, it is necessary to vary the number of shared properties between query and support set instances. The same image can be placed in different contexts in this way with a different result in the task of aspect-based FSL. The context of the support set is the crucial part in identifying the correct match by varying the shared properties. (see Figure \ref{fig:results_shapes} for example of support sets)

\subsubsection{Data split} 

In FSL the generalization of the model is evaluated on a validation set consisting of unseen classes. As we do not assume a set of pre-existing classes, we cannot employ such a split. In our controlled setting, we can design the dataset so that each image differs in at least one property. Therefore, each image can be classified as a separate class. We refer to this as the \textit{unique data split}. As a consequence, the training data effectively consists of classes that do not appear in the test data ($C_{training} \cap C_{test} = \emptyset$) as there are different images for training compared to testing.

Furthermore, we evaluate the method with a second approach, referred to as the \textit{aspect data split}. In this split, we use all the data points to form the support sets. However, we form unique sets of query images for training and testing. Therefore, the aspects that are present in training are not present in testing. The goal of this evaluation is not to evaluate the general purpose representation learning properties of the model, but rather to evaluate the capability of the model to correctly extract the aspect, or rather how well it generalizes in doing so. 
% Such an approach highlights two important parts. First, the perception of similarity differs based on the support set. 

% We can evaluate if the support set extracts the aspect by separating training- and testing query images. Especially when an image of the support set will always be seen as a negative example of a property during training. 

%Moreover, the testing queries have never been seen in the aspect-based embedding space for this support set during training. % I do not understand this. 

%So, the overlap in images in the support set for training and testing does not pose a problem.

\subsubsection{Loss} The tuplet loss \cite{sohn2016improved} generalizes the triplet loss \cite{schroff2015facenet} to explore multiple negative examples simultaneously. Instead of the triplet ($x, x^+, x^-$), we have a tuplet defined as ($x_a, x^+, x^-_1, .., x^-_{N-1}$), where N is the size of the support set. The tuplet loss (Equation \ref{eq:tuplet}) brings the anchor embedding to be closer to the positive example, as does the triplet. However, the loss pushes away all negative samples simultaneously instead of only one negative example.

\begin{equation}\label{eq:tuplet}
    L_{tuplet}(x, x^+, x^-_1, .., x^-_{N-1}) = \text{log}\bigg(1+\sum_{j=1}^{N-1}e^{\|f(x) - f(x^+)\|_2^2 - \|f(x) - f(x^-_j)\|_2^2}\bigg)
\end{equation}

\subsubsection{Evaluation metric}
% In aspect-based few shot learning, we formulate the matching task based on an aspect rather than a class. In contrast to the class, we define the aspect as a set of properties that are shared between the query and only one element of the support set. 

%For a given element $x_m$ part of the support set $\mathrm{S}$ ($x_m \in \mathrm{S}$). The query $x_q$ matches $x_m$ if there exists an aspect $\mathrm{A}$ such that $\mathrm{A} \subset x_m$ and $\mathrm{A} \subset x_q$ and  $\mathrm{A} \not \subset x_i$ for all $i$ where $i \not = m$.
%
%
%
As a metric, we use the Euclidean distance of our aspect-based FSL embedding versus the traditional FSL embedding. Specifically, we use the distance ratio\ref{eq:distance-ratio}.

%The accuracy is not the correct evaluation method for these experiments. The positive examples in our support sets always have one property more comparatively with the query than our negative examples. Therefore, a non-aspect-based embedding space would also capture the smallest distance. Rather, the distance between the query and positive examples and between the query and negative examples is a relevant evaluation method. 

For simplicity, we refer to the distance between the positive example and the query as the positive distance from now on. The same applies to the negative examples as the negative distance. The positive distance within the aspect-based embedding space is expected to be smaller than within a single embedding space due to the focus on the aspect rather than on all properties simultaneously. 

We also opt for a relative comparison: the ratio of positive distance to the negative distance. A larger ratio represents a better distinction between the positive and negative example:

\begin{equation*}
    \text{distance ratio} = \frac{|\text{average positive distance - average negative distance}|}{\text{average positive distance}}
    \label{eq:distance-ratio}
\end{equation*}
\subsubsection{Implementation details}
The experiments are implemented in PyTorch(version 2.1.0+cu121). For training, the learning rate is set to 7e-4. We use the ADAM optimizer with L2 regulization set to 1e-2. Each model trains for 50 epochs without any early stopping. The model is saved when the validation loss decreases compared to the best model up to this point. During testing, we generated ten different support sets for each query. Only one support set does not offer enough evidence as the support sets are generated randomly. Therefore, we generate multiple support sets.

\subsection{Results}
\begin{figure}[!b]
    \centering
    \includegraphics[scale=.35]{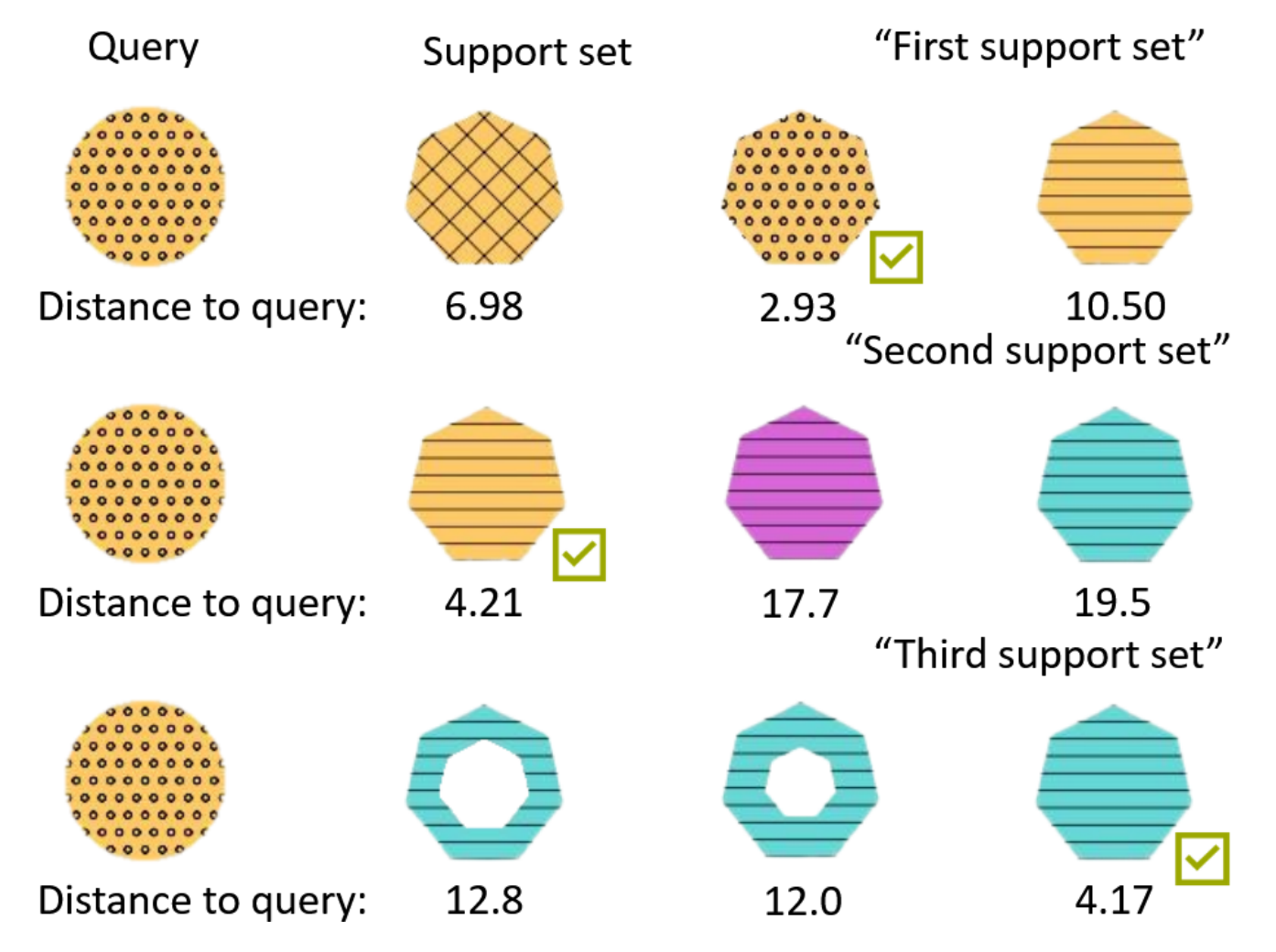}
    \caption{Example of the distance between the query and support set instances for the geometric shape data. Differentiating aspects leads to different distances between query and the same image.}
    \label{fig:results_shapes}
\end{figure}
\begin{table}[!b]
\centering
\resizebox{\textwidth}{!}{%
\begin{tabular}{l|ccc|ccc}
\toprule
\multirow{3}{*}{\textbf{Model}} & \multicolumn{3}{c|}{\textbf{Data Unique Split}} & \multicolumn{3}{c}{\textbf{Query Split}} \\ 
\cmidrule(lr){2-4} \cmidrule(lr){5-7} 
 & \textbf{Positive} & \textbf{Negative} & \textbf{Distance} & \textbf{Positive} & \textbf{Negative} & \textbf{Distance} \\ 
 & \textbf{Distance} & \textbf{Distance} & \textbf{Ratio} & \textbf{Distance} & \textbf{Distance} & \textbf{Ratio} \\ 
\midrule
\textbf{Baseline} & - & - & - & $49.0 \pm 0.46$ & $54.9 \pm 0.28$ & $0.1$-$0.14$ \\ 
\textbf{Shallow + Single Layer DSTM} & $5.46 \pm 0.08$ & $9.72 \pm 0.13$ & $0.73$-$0.84$ & $3.96 \pm 0.09$ & $7.57 \pm 0.14$ & $\underline{0.84}$-$\underline{0.99}$ \\ 
\textbf{VGG + Residual Block DSTM} & $3.23 \pm 0.08$ & $6.80 \pm 0.12$ & $\underline{1.02}$-$\underline{1.20}$ & $4.03 \pm 0.09$ & $8.14 \pm 0.15$ & $\mathbf{\underline{0.94}}$-$\mathbf{\underline{1.10}}$ \\ 
\textbf{ResNet + Residual Block DSTM} & $3.58 \pm 0.08$ & $8.85 \pm 0.18$ & $\mathbf{\underline{1.36}}$-$\mathbf{\underline{1.58}}$ & $4.59 \pm 0.09$ & $8.06 \pm 0.12$ & $0.69$-$0.81$ \\ 
\bottomrule
\end{tabular}%
}
\vspace{1em}

\caption{Geometric shapes data set: The average positive and negative distances of the second support set with a 95\% confidence interval. The distance ratio interval is calculated based on the 95\% confidence interval.}
\label{tab:support1_results}
\end{table}

\begin{table}[!t]
\centering
\resizebox{\textwidth}{!}{%
\begin{tabular}{l|ccc|ccc}
\toprule
\multirow{3}{*}{\textbf{Model}} & \multicolumn{3}{c|}{\textbf{Data Unique Split}} & \multicolumn{3}{c}{\textbf{Query Split}} \\ 
\cmidrule(lr){2-4} \cmidrule(lr){5-7} 
 & \textbf{Positive} & \textbf{Negative} & \textbf{Distance} & \textbf{Positive} & \textbf{Negative} & \textbf{Distance} \\ 
 & \textbf{Distance} & \textbf{Distance} & \textbf{Ratio} & \textbf{Distance} & \textbf{Distance} & \textbf{Ratio} \\ 
 \midrule
\textbf{Baseline} & - & - & - & $40.5 \pm 0.56$ & $49.0 \pm 0.32$ & $0.19$-$0.24$ \\ 
\textbf{Shallow + Single Layer DSTM} & $4.03 \pm 0.07$ & $8.96 \pm 0.14$ & $1.15$-$1.30$ & $2.66 \pm 0.06$ & $6.67 \pm 0.14$ & $\underline{1.40}$-$\underline{1.63}$ \\ 
\textbf{VGG + Residual Block DSTM} & $2.26 \pm 0.06$ & $6.19 \pm 0.13$ & $\underline{1.60}$-$\underline{1.85}$ & $2.95 \pm 0.07$ & $7.84 \pm 0.16$ & $\mathbf{\underline{1.54}}$-$\mathbf{\underline{1.79}}$ \\ 
\textbf{ResNet + Residual Block DSTM} & $2.54 \pm 0.06$ & $8.32 \pm 0.19$ & $\mathbf{\underline{2.13}}$-$\mathbf{\underline{2.42}}$ & $3.24 \pm 0.07$ & $7.25 \pm 0.12$ & $1.15$-$1.32$ \\ 
\bottomrule
\end{tabular}%
}
\vspace{1em}
\caption{Geometric shapes data set: The average positive and negative distance of the second support set with a 95\% interval. The distance ratio interval is calculated based on the 95\% interval.}
\label{tab:support2_results}
\end{table}
We provide three options for representation learning: a shallow representation model, a small VGG architecture, and a small ResNet architecture. The shallow representation model is a simple architecture of convolutional layer + batch normalization + ReLU followed by a pooling layer. The baseline model would be the shallow representation model without the DSTM. Without the DSTM, we can see the effect of the module on the embedding space. Within the DSTM, we experiment with two types of layers in the permutation-equivariant and -variant deep-set models. The first option is to have a single convolutional layer + batch normalization + ReLU for the embedding function $g(;, \mu)$ in the permutation-equivariant model and a single layer for the embedding function $k(;\eta)$ in the permutation-invariant model. The second option is to have a single residual block in those places in the model instead.
\subsubsection{Analysis of the distance ratio} Figure \ref{fig:results_shapes} shows the three different support sets for the same query. The upper support set is referred to as the first support set. This support set has two shared features with the query. In this example, the shared features are color and thickness. In the first support set, we see that the second image matches the query on the aspect of the pattern. The third image of the first support set has a distance of 10.5 from the query image. However, we see that the same image is much closer to the query in the middle or second support set, given the aspect of color. This second support set has only one shared feature in thickness. The distance from the query depends on the aspect of the support set that aligns with the goal of this work. We see the same happening with the third image of the second support set transferred to the lower or third support set. In this support set, we do not have any shared features.

Table \ref{tab:support1_results} shows the average positive distance and negative distance with a 95\% interval of the first support set for the best-performing model configuration and the baseline. The boundaries of the distance ratio are calculated on the basis of the interval 95\%. We see that the distance ratio of the baseline is small. So, the baseline model cannot differentiate between positive and negative examples well. Each model trained with the DSTM has a better distance ratio. So, the model does a better job of differentiating between positive and negative examples. The VGG and ResNet representation models with the residual block DSTM perform best for the unique split of the data. The VGG representation model with the residual block DSTM also performs well in the case of a query split. However, the shallow representation model with a single-layer DSTM performs better than the ResNet representation model with the residual block DSTM.

\begin{table}[!b]
\centering
\resizebox{\textwidth}{!}{%
\begin{tabular}{l|ccc|ccc}
\toprule
\multirow{3}{*}{\textbf{Model}} & \multicolumn{3}{c|}{\textbf{Data Unique Split}} & \multicolumn{3}{c}{\textbf{Query Split}} \\ 
\cmidrule(lr){2-4} \cmidrule(lr){5-7} 
 & \textbf{Positive} & \textbf{Negative} & \textbf{Distance} & \textbf{Positive} & \textbf{Negative} & \textbf{Distance} \\ 
 & \textbf{Distance} & \textbf{Distance} & \textbf{Ratio} & \textbf{Distance} & \textbf{Distance} & \textbf{Ratio} \\ 
 \midrule
\textbf{Baseline} & - & - & - & $14.2 \pm 0.05$ & $34.6 \pm 0.1$ & $1.42$-$1.45$ \\ 
\textbf{Shallow + Single Layer DSTM} & $1.24 \pm 0.01$ & $7.73 \pm 0.03$ & $5.16$-$5.31$ & $1.19 \pm 0.01$ & $10.8 \pm 0.05$ & $8.01$-$8.24$ \\ 
\textbf{VGG + Residual Block DSTM} & $0.91 \pm 0.01$ & $6.67 \pm 0.04$ & $\mathbf{6.20}$-$\mathbf{6.37}$ & $0.76 \pm 0.01$ & $10.1 \pm 0.04$ & $\mathbf{12.0}$-$\mathbf{12.46}$ \\ 
\textbf{ResNet + Residual Block DSTM} & $1.03 \pm 0.01$ & $6.75 \pm 0.05$ & $5.39$-$5.60$ & $1.05 \pm 0.01$ & $11.3 \pm 0.04$ & $9.58$-$9.86$ \\ 
\bottomrule
\end{tabular}%
}

\vspace{1em}

\caption{Sprites data set: The average positive and negative distances of the first support set with a 95\% confidence interval. The distance ratio interval is calculated based on the 95\% interval.}
\label{tab:results_sprites1}
\end{table}

Table \ref{tab:support2_results} displays the average positive distance and negative distance with a 95\% interval of the second support set for each model configuration and the baseline. The distance ratios are smaller than the first support set in Table \ref{tab:support2_results}. There is an increase in the positive distance and the negative distance. It indicates that the DSTM cannot completely filter out the differentiating features between the support set and the query. However, the best model architectures can still differentiate relatively well between negative and positive examples.

Table \ref{tab:results_sprites1} displays the average positive distance and negative distance with a 95\% interval of the first support set for each model configuration and the baseline. The distance ratio of the first support set displays the best results in the case of the query data split. Each DSTM architecture with a query split performs better than every model with a unique data split. However, the unique data split still outperforms the baseline by a large margin. The distance ratios in Table \ref{tab:results_sprites1} are also an improvement compared to the results of the geometric shapes data set in Table \ref{tab:support1_results}.

\section{Conclusion}
This work addressed the intricate problem of label-based supervision approaches in Few-Shot learning. Traditional Few-Shot learning approaches assume that a single class describes the
content of a data point. The class label of the query must match precisely one of the support set class labels. We generalized this class-based assignment by introducing the notion of an aspect. 

%It does not reflect the human understanding of the data. We do not consider only the images separately. The context of the presented support set also conveys information to us. The different and shared information among support set instances matters in our decision-making. In this work, we proposed to extend the Few-shot methods. 

The experimental results with the DSTM show an improved differentiation between positive and negative examples when we do not consider a single-class assignment to the images. However, this evaluation is still limited to the synthetic cases where we can fully control the properties of the data to form the aspect. Although this is a good initial step in assessing the validity of this approach, to demonstrate the real value of this approach, a user study with natural image data needs to be implemented where the capacity of DSTM to infer the aspect can be contrasted with the innate ability of humans to do the same. 


\bibliographystyle{splncs04}
\bibliography{mybibliography}
%Author, F.: Article title. Journal \textbf{2}(5), 99--110 (2016)
%Author, F., Author, S.: Title of a proceedings paper. In: Editor,
%F., Editor, S. (eds.) CONFERENCE 2016, LNCS, vol. 9999, pp. 1--13.
%Springer, Heidelberg (2016). \doi{10.10007/1234567890}
%Author, F., Author, S., Author, T.: Book title. 2nd edn. Publisher,
%Author, A.-B.: Contribution title. In: 9th International Proceedings
%on Proceedings, pp. 1--2. Publisher, Location (2010)
%LNCS Homepage, \url{http://www.springer.com/lncs}, last accessed 2023/10/25
%\end{thebibliography}
\end{document}